# Decayed MCMC Filtering

**Bhaskara Marthi, Hanna Pasula, Stuart Russell, Yuval Peres**
University of California, Berkeley
Berkeley, CA 94720
{bhaskara, pasula, russell}@cs.berkeley.edu
peres@stat.berkeley.edu

## Abstract

Filtering—estimating the state of a partially observable Markov process from a sequence of observations—is one of the most widely studied problems in control theory, AI, and computational statistics. Exact computation of the posterior distribution is generally intractable for large discrete systems and for nonlinear continuous systems, so a good deal of effort has gone into developing robust approximation algorithms. This paper describes a simple stochastic approximation algorithm for filtering called *decayed MCMC*. The algorithm applies Markov chain Monte Carlo sampling to the space of state trajectories using a proposal distribution that favours flips of more recent state variables. The formal analysis of the algorithm involves a generalization of standard coupling arguments for MCMC convergence. We prove that for any ergodic underlying Markov process, the convergence time of decayed MCMC with inverse-polynomial decay remains bounded as the length of the observation sequence grows. We show experimentally that decayed MCMC is at least competitive with other approximation algorithms such as particle filtering.

## 1 Introduction

Let us consider a partially observable Markov process with state variable $X_t$ and observation variable $Y_t$. The process is described by a transition model $P(X_{t+1}|X_t)$, a sensor model $P(Y_t|X_t)$, and a prior $P(X_0)$ (see Figure 1). The process is assumed to be *stationary*—the transition and sensor models do not vary with $t$—and ergodic. At any given current time $T$, the observations $y_1, \ldots, y_T$ (abbreviated as $y_{1:T}$) are available. The basic problem of calculating $P(X_T|y_{1:T})$—the *belief state* or distribution over possible states given the evidence to date—has been studied in many guises, as state estimation, filtering, tracking, or situation assessment. We will use the term "filtering," and we will concentrate on two aspects: (1) the *update* computation needed when a single new observation arrives, and (2) the behaviour of the filtering algorithm in the limit of long observation sequences (i.e., as $T \to \infty$).

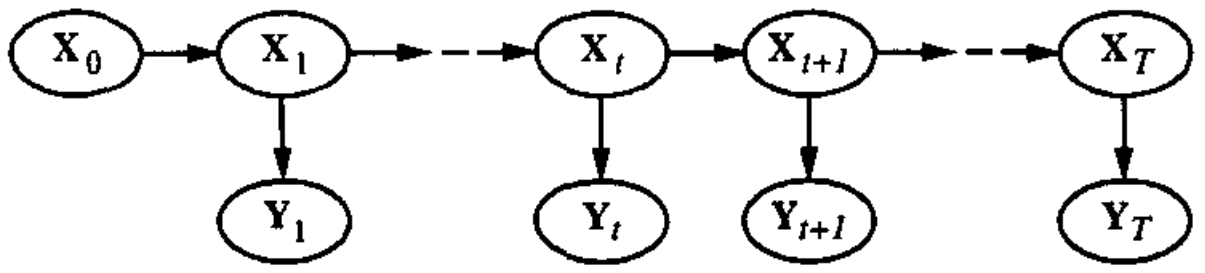

Figure 1: Unrolled Bayesian network depicting a partially observable Markov process.

Markov processes come in various flavours: discrete models such as hidden Markov models (HMMs) and discrete dynamic Bayesian networks (DBNs); continuous models such as Kalman filters; and hybrid models such as switching Kalman filters. All of these approaches are expressible as generalized DBNs with $X_t = X_t^1, \ldots, X_t^n$ and $Y_t = Y_t^1, \ldots, Y_t^m$. Exact update is intractable for several of the standard classes—all existing algorithms are $O(2^n)$ for discrete DBNs and $O(\infty)$ for switching Kalman filters (i.e., the update cost grows without bound as $T \to \infty$). A variety of approximation algorithms have therefore been suggested, and several will be discussed in Section 2. Particle filtering (PF) in particular is a robust and general algorithm with many applications [Doucet *et al.*, 2001]. Four crucial features of PF are (1) it takes a constant amount of time per update, independent of $T$ (which is essential for any online filtering algorithm); (2) by increasing the number of samples it can approach the true belief state arbitrarily closely; (3) it can be applied easily to any standard Markov process model; and (4) it is usually *non-divergent*—i.e., its estimation error remains bounded for large $T$. There are cases, however, where particle filtering diverges. Particle filtering also has the drawback that its space requirement is proportional to the number of samples used.

The main contribution of this paper is a new filtering algorithm called *decayed Markov chain Monte Carlo*, or decayed MCMC (Section 3). The basic idea is to concentrate the sampling activity of the MCMC algorithm on state variables in the recent past, since they are more relevant to the current state. Decayed MCMC shares the advantages of particle filtering but is provably convergent given certain standard conditions on the Markov process being observed. In Section 4 we develop a generalized form of the standard *coupling lemma* used to analyze convergence of MCMC algorithms, and we prove that a particular form of decayed MCMC using an inverse polynomial decay converges to within an arbitrary $\epsilon$ of the true belief state in time that is independent of $T$, the length of the observation sequence. This implies that decayed MCMC is non-divergent. In Section 5, we demonstrate empirically that our algorithm's performance is comparable to that of PF. We draw our conclusions in Section 6.

## 2 Approximate DBN inference methods

One reason why exact DBN inference is intractable is that the running time of BN algorithms is exponential in the tree width of the underlying graph. In a DBN, the existing dependencies will cause this quantity to grow to $n$ as the network is unrolled. Boyen and Koller [1998] have suggested that this problem can be overcome by, at every timestep, ignoring some of the weaker variable interdependecies. This approach has been shown to work very well on some DBNs. The downside is that picking the dependencies to be ignored is a non-trivial problem, difficult to automate. Moreover, once the simplifications have been selected, the error is a deterministic function of the graph and of the set of observations, so it is not possible to make arbitrarily close approximations.

An alternative method is Loopy Belief Propagation [Murphy and Weiss, 2001]. Here, belief propagation, which is a tractable exact algorithm for polytree BNs, is applied to an arbitrary DBN until convergence. This approach gives approximate answers, but there are no guarantees as to their quality; once again, no arbitrary improvement of the approximation is possible. Recent generalizations of belief propagation [Yedidia *et al.*, 2001; Minka, 2001] do admit of successively more accurate approximations and may yield a practical filtering algorithm.

A third deterministic approximation algorithm can be derived using variational techniques, which use the "closest" simplified model that is tractable. The original variational algorithms were derived for specific families of DBN structures, e.g., factorial HMMs [Ghahramani and Jordan, 1997], and resemble the Boyen–Koller algorithm in that they perform well if the variational model is a good fit but cannot produce arbitrarily close approximations.

Particle filtering [Doucet *et al.*, 2001], the most widely applied algorithm, represents the belief state by a set of samples. The samples are propagated forward at every time step, weighted according to the likelihood of the new observation, and then resampled according to the weights so as to move the sample population towards the high-likelihood part of the state space. As $S$, the number of samples, is increased, the approximation becomes arbitrarily good. Formal analysis of convergence has proven quite difficult, and the basic algorithm can diverge when the diversity of the sample population collapses.

## 3 Decayed MCMC

### 3.1 MCMC filtering

MCMC [Gilks *et al.*, 1996] generates samples from a posterior distribution $\pi(x)$ over possible worlds $x$ by simulating a Markov chain[1] whose states are the worlds $x$ and whose *stationary distribution* is $\pi(x)$. Even though the samples are not independent, the ergodic theorem guarantees that expectations estimated from the samples converge to the right answer as $S \to \infty$.

For filtering, it would be natural to construct a Markov chain such that the posterior distribution $\pi(x)$ is the belief state $P(X_T|y_{1:T})$. Unfortunately, there is no satisfactory way to define this chain without considering the values of $X_{1:T-1}$ as well, and so the target stationary distribution $\pi$ will be $P(X_{1:T}|y_{1:T})$. This means that the sample worlds visited by the computational Markov chain will be complete state trajectories over $X_{1:T}$. The estimated belief state $P(X_T|y_{1:T})$ can be extracted easily from the sampled trajectories simply by looking at the value of $X_T$ in each trajectory. The memory requirements of MCMC filtering are therefore independent of the number of samples (and therefore, of the required accuracy), unlike particle filtering, since the algorithm can simply accumulate counts for each value of $X_T$ (or for particular $X_T^i$ variables that may be queried). On the other hand, the MCMC algorithm samples past states conditioned on their Markov blankets, so we must store the history of evidence $\{y_t\}$. In practice, there will be a limit $L$ such that evidence more than $L$ steps in the past is forgotten. However, the convergence time of the algorithm does not depend on $L$, and so a pessimistic value of $L$ can be used without affecting performance.

A computational Markov Chain with the appropriate stationary distribution can be constructed using *Gibbs sampling* [Pearl, 1988]. Viewing the model as having a single state variable $X_t$, the Gibbs sampling step first chooses $t$ and then samples $X_t$ from the distribution conditioned on its Markov blanket,

$$P(X_t|X_{t-1}, X_{t+1}, Y_t) = \alpha P(X_t|X_{t-1})P(Y_t|X_t)P(X_{t+1}|X_t).$$

(With multiple state variables, each state variable $X_t^i$ is sampled conditional on its own Markov blanket.) Gibbs sampling is local in that the Markov blanket involves nodes in a neighbourhood of $X_t^i$; each sampling step takes time that is, for bounded fan-in, independent of the model size.

[1] This "computational" Markov chain should be distinguished from the "physical" Markov chain whose state is being estimated.

With Gibbs sampling, the order in which the $X_t$'s are suggested as candidates for change can be fixed. It is also permissible to pick variables at random from some distribution $g(t)$ over $[1\ldots T]$, as long as every variable is guaranteed to be chosen infinitely often. The algorithm (for the single state-variable case) is as follows:

```
for s=1 to S
  choose t from g(t)
  sample X_t from P(X_t|X_{t-1}, X_{t+1}, Y_t)
  update counts for X_T
```

For reasons that will become clear, the choice of the *decay function* $g(t)$ is crucial to the success of MCMC filtering.

### 3.2 The decay function

We are concerned primarily with the *mixing time* $\tau(\epsilon)$ of the MCMC process, which, roughly speaking, is the number of samples required before the estimated posterior is within an error tolerance $\epsilon$ of the true posterior. (A more precise definition is given in Section 4.) In most analyses of MCMC, one measures error with respect to the posterior over states of the computational Markov process. In our case, that would mean $P(X_{1:T}|y_{1:T})$, the posterior over trajectories. For filtering, however, we are interested only in the error with respect to the posterior marginal $P(X_T|y_{1:T})$. Let us consider a number of possible choices for the decay function $g(t)$ and see how the choice affects the mixing time.

**Uniform over** $[1\ldots T]$:
$g_T(t)=1/T$ for $1\leq t\leq T$, 0 otherwise.
This is the usual way to apply Gibbs sampling to Bayesian networks, with every variable sampled equally often. For the posterior marginal error at $X_T$ to be less than $\epsilon$, we must sample $X_T$ some number of times proportional to some increasing function of $1/\epsilon$, and the total amount of work will be $T$ times larger than this. Therefore, MCMC filtering with a uniform decay function fails as $T\to\infty$, because the cost per update grows without bound.

A uniform decay fails because it spends arbitrary amounts of time sampling variables in the far distant past that are essentially irrelevant to the current state. More precisely, if the "physical" Markov process (conditioned on the evidence) is ergodic, old values of both the observations and the states are forgotten exponentially fast with a rate that can be bounded by the Birkhoff coefficients of the process [Shue *et al.*, 1998]. Thus, it is helpful to think of a *physical mixing time* $\tau_p$ for the observed process.

**Uniform over fixed window** $[(T-W+1)\ldots T]$ :
$g_W(t)=1/W$ for $(T-W+1)\leq t\leq T$, 0 otherwise.
Uniform sampling over the recent past has the advantage that the marginal at $X_T$ will converge in time that depends only on the window size $W$ and not on $T$; it has the disadvantage that it converges to the wrong distribution unless $W$ is chosen to be much larger than the physical mixing time $\tau_p$ (which is typically unknown). Further, once $W$ has been fixed, arbitrary improvements in the accuracy cannot be made. Finally, the fixed-window approach spends as much time flipping variables at time $T-W+1$ as it does variables at time $T$, which is wasteful.

**Exponential decay**:
$g_\beta(t)=\alpha_\beta e^{-\beta(T-t)}$ for $1\leq t\leq T$, 0 otherwise.
Since an exponential decay ensures that every $t$ is sampled infinitely often in the limit, convergence to the correct marginal at $X_T$ is guaranteed. If the decay constant $\tau_g=1/\beta$ is matched to the physical mixing time $\tau_p$, we expect reasonably fast convergence because the sampling frequency is proportional to "relevance." However, since $\tau_p$ is unknown, there is a danger of setting $\tau_g$ too large (in which case samples in the far past are wasted) or too small (in which case the number of samples needed for convergence to the correct marginal grows exponentially in the difference $\tau_p-\tau_g$ and also with $1/\epsilon$).

**Inverse polynomial decay**:
$g_\delta(t)=\alpha_\delta(T-t+1)^{-(1+\delta)}$ for $1\leq t\leq T$, 0 otherwise.
Again, we have convergence to the correct marginal in the limit. We prove in Section 4 that the inverse polynomial decay results in a convergence time that is independent of $T$. Moreover, because the proof does not depend on the starting state of the MCMC algorithm, decayed MCMC is robust against divergence as $T\to\infty$.

## 4 A mixing time bound

We now prove a bound on the mixing time of decayed MCMC with an inverse polynomial decay $g_\delta(t)$ for discrete DBNs. The bound does not depend on the history length. For simplicity, we assume the DBN has one state variable and one observation variable (this will not affect the asymptotic behaviour of the mixing time).

### 4.1 Notation

We begin by introducing some notation. All discussion of mixing times is assumed to be with respect to some prespecified $\epsilon$. The *total variation distance* between two probability distributions on a set $S$ is defined as

$$||p_1-p_2||=\frac{1}{2}\sum_{x\in S}|p_1(x)-p_2(x)|$$

The state and observation variables of the DBN take values in the finite sets $\mathcal{X}$ and $\mathcal{Y}$ respectively. $T$ will denote the length of the evidence sequence. We define the *mixing parameter* $\eta$ of a DBN as the maximum, over all values $x_{t-1},x_{t+1},x'_{t-1},x'_{t+1}\in\mathcal{X}$ and $y_t\in\mathcal{Y}$, of

$$||P(X_t|x_{t-1},x_{t+1},y_t)-P(X_t|x'_{t-1},x'_{t+1},y_t)||$$

$\eta$ will be part of the constant factor in our mixing time analysis, and it summarizes the mixing properties of the DBN. For a given evidence sequence $y$, a tighter "data-dependent" version of $\eta$ can be used by not maximizing over $y_t$.

Our MCMC notation is from Jerrum and Sinclair [1997]. The state space of the computational MCMC process is $\Omega = \mathcal{X}^T$, the set of all physical trajectories of length $T$. The stationary distribution of MCMC on $\Omega$ is denoted by $\pi$. $P_x^s$ will denote the probability distribution on $\Omega$ resulting from starting in state $x \in \Omega$, and running MCMC for $s$ steps. In general, superscripts will refer to the number of time steps of MCMC, and subscripts to time steps in the DBN, so that $X_t^s$ is the state of the $t^{th}$ timeslice of the DBN after $s$ steps of MCMC. (This conflicts slightly with the earlier use of superscripts as identifiers of individual variables within a timeslice, but we will avoid the latter usage in what follows.) $\Delta^s(x)$ denotes the *error*—the total variation distance between the MCMC distribution at step $s$ and the stationary distribution, i.e., $||P_x^s - \pi||$. The worst-case distance for all starting states is $\Delta^s = \max_x \Delta^s(x)$. The *mixing time* $\tau(\epsilon)$ is then $\min\{s|\Delta^s < \epsilon\}$, i.e., the first time at which the worst-case distance is less than $\epsilon$. We will often omit the dependence on $\epsilon$, since it is a prespecified constant.

We are specifically interested in the $T^{th}$ timeslice, and so define $\mathcal{M}$ to be the operator that takes a probability distribution on $\Omega$ and marginalizes it onto the last coordinate. We can then define the marginal error $\Delta_m^s(x) = ||\mathcal{M}(P_x^s) - \mathcal{M}(\pi)||$, and use this to define $\Delta_m^s$ and $\tau_m(\epsilon)$ as before. The marginal mixing time $\tau_m$ is the quantity we want to bound.

### 4.2 Coupling and Marginal Coupling

The technique of coupling [Bubley and Dyer, 1997] is commonly used in proving bounds on the mixing time of MCMC algorithms. The idea is that we consider two instances of the chain, and bound the mixing time of the chain in terms of how long the two instances take to come together. Now, if the two instances were independent, this would not be a very useful thing to do because the bound would be very loose. However, the power of the method is that we may "couple" the two instances together however we like, by specifying their joint transition matrix, so long as their marginal transition behaviour is according to the given Markov chain, and the coupling bounds will still hold. More precisely, we have the following theorem :

**Theorem 1 (MCMC Coupling Theorem)** *Given a Markov transition matrix K, let $\{X^s\}$ and $\{\tilde{X}^s\}$ be two Markov chains such that*

- *For each s, the marginal transitions $P(X^{s+1}|X^s)$ and $P(\tilde{X}^{s+1}|\tilde{X}^s)$ are given by K*
- $X^s = \tilde{X}^s \Rightarrow X^{s+1} = \tilde{X}^{s+1}$

*Then the mixing time satisfies $\tau(\epsilon) < \bar{S}/\epsilon$, where*

$$\bar{S} = \max_{x,\tilde{x}} E(\min\{s|X^s = \tilde{X}^s\}|X^0 = x, \tilde{X}^0 = \tilde{x}) .$$

We cannot use this theorem directly, because we want to bound the marginal mixing time $\tau_m$ rather than the mixing time $\tau$ for the entire sequence. Of course, $\tau_m \leq \tau$, but because $\tau$ depends on $T$, this bound is too weak for our purposes. Therefore, we prove a modified version of the coupling theorem. First, we recall a lemma from probability theory.

**Lemma 1 (Coupling Lemma)** *Let U and V be discrete random variables with distributions given by f and g. Then*

1. $P[U \neq V] \geq ||f - g||$
2. *There exists a joint distribution for U and V with marginals f and g that allows equality to be achieved in the above.*

Now let $D_m^s(x, \tilde{x})$ be the marginal distance after $s$ steps between two MCMC processes starting from states $x$, $\tilde{x}$, i.e., $D_m^s(x,\tilde{x}) = ||\mathcal{M}(P_x^s) - \mathcal{M}(P_{\tilde{x}}^s)||$. As before, we will be concerned with the worst-case marginal distance: $D_m^s = \max_{x,\tilde{x}} D_m^s(x,\tilde{x})$. We can show that this gives an upper bound on the marginal error:

**Lemma 2** $\Delta_m^s \leq D_m^s$.

**Proof:** For $x \in \Omega$, let $P^0(x)$ be the probability distribution that assigns 1 to $x$ and 0 to anything else. We can then write $\pi$ as a convex combination $\sum_{x\in\Omega} a_x P^0(x)$ where $a_x \geq 0$ and $\sum_x a_x = 1$.

We can view a probability distribution $p$ over $\Omega$ as a vector, and the transition kernel of MCMC as a matrix $K$, so that if we apply one step of MCMC to a distribution $p$, we obtain the distribution $pK$. Since $\pi$ is stationary,

$$\begin{aligned} \pi &= \pi K^s \\ &= \sum_x a_x P^0(x) K^s = \sum_x a_x P_x^s \end{aligned}$$

Since $\mathcal{M}$ is also linear, $\mathcal{M}(\pi) = \sum_x a_x \mathcal{M}(P_x^s)$, i.e., $\mathcal{M}(\pi)$ is contained in the convex hull of the $\mathcal{M}(P_x^s)$.

Let $\tilde{x} \in \Omega$, and consider the ball centered at $\mathcal{M}(P_{\tilde{x}}^s)$ with radius $\max_x ||\mathcal{M}(P_{\tilde{x}}^s) - \mathcal{M}(P_x^s)||$. Since this is convex and contains all the $\mathcal{M}(P_x^s)$, it must also contain $\mathcal{M}(\pi)$, and so, for any $\tilde{x}$,

$$\begin{aligned} ||\mathcal{M}(P^s(\tilde{x})) - \mathcal{M}(\pi)|| &\leq \max_x ||\mathcal{M}(P^s(\tilde{x})) - \mathcal{M}(P^s(x))|| \\ &\leq D_m^s \end{aligned}$$

The claim follows by taking a maximum over $\tilde{x}$. ∎

We will use these lemmas to prove a marginal version of Theorem 1. Essentially, instead of looking at the time it takes until $X^s = \tilde{X}^s$, we just look at how long it takes until they agree on their $T^{th}$ coordinate, i.e., $X_T^s = \tilde{X}_T^s$. Now, we can no longer require that the chains stay together once they come together on the $T^{th}$ coordinate, because that would violate the requirement that each chain's dynamics mirror the specified Markov chain. However, we can still get a bound on total variation distance after $S$ steps. In our applications, this bound will be a non-increasing function of $S$, and so we get a bound on $\tau_m$ as well.

**Theorem 2 (Marginal Coupling Theorem)** *For a given transition matrix $K$ on $\Omega$, let $\{X^s\}$ and $\{\tilde{X}^s\}$ be Markov chains such that for all $s$, $P(X^{s+1}|X^s)$ and $P(\tilde{X}^{s+1}|\tilde{X}^s)$ are given by $K$. Then*

$$\Delta_m^s \leq \max_{x,\tilde{x}} P[X_T^s \neq \tilde{X}_T^s | X^0 = x, \tilde{X}^0 = \tilde{x}]$$

**Proof:** By Lemma 2, $\Delta_m^s \leq D_m^s$. Since $\{X^s\}$ and $\{\tilde{X}^s\}$ both evolve marginally according to $K$, $D_m^s(x, \tilde{x})$ equals

$$||P(X^s|(X^0, \tilde{X}^0) = (x, \tilde{x})) - P(\tilde{X}^s|(X^0, \tilde{X}^0) = (x, \tilde{x}))||$$

We can now apply part 1 of Lemma 1 to finish the proof. ■

Another useful extension of the coupling framework is *multiple-step coupling*. Suppose we have a Markov chain with dynamics given by the transition matrix $K$, and we want to show that the mixing time is less than $S$. To use the coupling theorem directly requires finding a coupling on a single step of $K$ which brings two instances together in $S$ steps with high probability. Sometimes, however, it is simpler to consider the $S$-step dynamics with transition matrix $K^S$, and find a coupling for this new dynamics that brings two instances together in 1 step with high probability. Since both $K$ and $K^S$ have the same stationary distribution, the existence of such a coupling would also imply that $K$ mixes in $S$ steps. This idea extends to marginal coupling, resulting in the following corollary to Theorem 2.

**Corollary 1** *Let $K$ be a transition matrix on $\Omega$, and $S > 0$. Suppose we can construct a coupling $(X, \tilde{X}) \to (X^S, \tilde{X}^S)$ such that $P(X^S|X)$ and $P(\tilde{X}^S|\tilde{X})$ are both given by $K^S$, and $P(X_T^S \neq \tilde{X}_T^S | X = x, \tilde{X} = \tilde{x}) < \epsilon\ \forall x, \tilde{x}$. Then the marginal mixing time of $K$ satisfies $\tau_m(\epsilon) < S$.*

### 4.3 The decayed window dynamics

Suppose that we have a polynomial decay function $g_\delta(t)$ but modify the Gibbs sampling algorithm so that it does nothing whenever $g_\delta$ chooses a time $t < T - W + 1$ for some fixed $W$. We call this the *decayed window* dynamics; its transition kernel is $K_{\delta,W}$. Since it ignores evidence before $T - W + 1$, its stationary distribution $\pi_W$ will not in general equal $\pi$. In this section, we will find a bound on the mixing time of the decayed window dynamics which depends on $W$ but not $T$. This result will then be used to bound the mixing time of decayed MCMC.

Given a matrix $K$ and vector $\phi$, define the *Dirichlet Form*

$$\mathcal{E}_{K,\phi}(f, f) = \frac{1}{2} \sum_{x,\tilde{x}} (f(x) - f(\tilde{x}))^2 K(x, \tilde{x}) \phi(x)$$

Let $\mathcal{F}_\phi$ be the family of nonnegative real-valued functions on $\Omega$ such that $\sum_x \phi(x) f^2(x) = 1$. For $f \in \mathcal{F}_\phi$, define the entropy $H_\phi(f^2) = \sum_x \phi(x) f^2(x) \log f^2(x)$. Finally, define the *logarithmic Sobolev constant* by

$$c_s(K, \phi) = \inf_{f \in \mathcal{F}_\phi} \frac{\mathcal{E}_{K,\phi}(f, f)}{H_\phi(f^2)}$$

The logarithmic Sobolev constant provides a bound on the mixing time, via the following theorem[2] [Diaconis and Saloff-Coste, 1996], [Randall and Tetali, 2000].

**Theorem 3** *For a Markov chain with transition kernel $K$ and stationary distribution $\pi$, with $\pi^* = \min_x \pi(x)$,*

$$\tau(\epsilon) < c_s^{-1}(K, \pi) \log(\log(1/\pi^*)) \log(1/\epsilon)$$

Define a matrix $K_{\text{unif},W}$ as follows : if $x, \tilde{x}$ differ only at the $t^{th}$ timeslice for some $t > T - W$, then $K_{\text{unif},W}(x, \tilde{x}) = P(X_t = \tilde{x}_t | X_{\text{mb}(t)} = x_{\text{mb}(t)})$; otherwise, $K_{\text{unif},W}(x, \tilde{x}) = 0$. In statistical physics, $K_{\text{unif},W}$ is an example of a generator of the Glauber dynamics of a lattice spin system. Its log Sobolev constant is bounded as follows: [Martinelli, 1999][3]

**Theorem 4** $c_s(K_{unif,W}, \pi_W) \geq C_1(\eta) > 0$ *where $C_1(\eta)$ is independent of $T$.*

$K_{\text{unif},W}$ is closely related to $K_{\delta,W}$, and we can use Theorem 4 to bound the mixing time of $K_{\delta,W}$.

**Theorem 5** *Given a DBN with mixing parameter $\eta$, the decayed window dynamics with window size $W$ and a polynomial decay function $g_\delta$ mixes to within $\epsilon/6$ of $\pi_W$ in $C_2(\eta, \delta) W^{1+\delta} \log(W) \log(1/\epsilon))$ steps where $C_2(\eta, \delta)$ is independent of $T$.*

**Proof:** Since $\min_{\{t > T-W\}} g_\delta(t) = g_\delta(T - W + 1)$, the Dirichlet forms of $K_{\text{unif},W}$ and $K_{\delta,W}$ satisfy the inequality

$$\mathcal{E}_{K_{\delta,W}, \pi_W}(f, f) \geq g_\delta(T - W + 1) \mathcal{E}_{K_{\text{unif},W}, \pi_W}(f, f)$$

$\mathcal{E}$ is the only thing in the definition of $c_s$ which depends on $K$. So, by Theorem 4,

$$\begin{aligned} c_s^{-1}(K_{\delta,W}, \pi_W) &\leq g_\delta^{-1}(T - W + 1) c_s^{-1}(K_{\text{unif},W}, \pi_W) \\ &= O(W^{1+\delta} C_1^{-1}(\eta)) \end{aligned}$$

Also, $\pi_W^* > C_3^{-W}$ for some constant $C_3$, and so $\log(\log(1/\pi_W^*)) = O(\log(W))$. Plugging all this into Theorem 3 gives the desired bound. ■

This bound on the mixing time implies the existence of a multiple-step coupling that makes two instances of the decayed window dynamics come together quickly:

[2]This bound is similar to classical bounds based on the eigengap of $K$, the main difference being that we have a term $\log(\log(1/\pi^*))$ instead of $\log(1/\pi^*)$, which turns out to be a crucial improvement

[3]See Theorem 4.6 of this reference, but note that the definition of $c_s$ used there is the inverse of our definition

**Corollary 2** *For $S \geq C_2(\eta)W^{1+\delta}\log(W)\log(1/\epsilon)$, there exists a coupling $(X,\tilde{X}) \to (X^S,\tilde{X}^S)$ such that if $P(X^S|X)$ and $P(\tilde{X}^S|\tilde{X})$ are given by $K^S_{\delta,W}$, then $\forall x,\tilde{x}\, P(X^S \neq \tilde{X}^S|X=x,\tilde{X}=\tilde{x}) \leq \epsilon/3$.*

**Proof:** By Theorem 5, the distributions $P(X^S|X=x)$ and $P(\tilde{X}^S|\tilde{X}=\tilde{x})$ are within $\epsilon/6$ of $\pi_W$. Therefore, by the triangle inequality, they are within $\epsilon/3$ of each other. By part 2 of Lemma 1, we can couple the chains so that they are equal with probability at least $1-\epsilon/3$. ■

### 4.4 Constructing a coupling

Let $K_\delta$ denote the decayed MCMC dynamics with inverse polynomial decay $g_\delta(t)$. We want to bound the mixing time using Corollary 1. To do this, we need to find a constant $S$, and, for all $x,\tilde{x} \in \Omega$, a coupling $P(X^S,\tilde{X}^S|X=x,\tilde{X}=\tilde{x})$ with the appropriate marginals, such that $P(X^S_T=\tilde{X}^S_T|X=x,\tilde{X}=\tilde{x}) > 1-\epsilon$.

Our strategy will be to couple the evolution $X \to X^S$ to an instance of the decayed window dynamics $X \to X^*$ with $P(X^*|X)$ given by $K^S_{\delta,W}$, and similarly couple the evolution $\tilde{X} \to \tilde{X}^S$ to $\tilde{X} \to \tilde{X}^*$. By the results of the previous section, we can choose $S=O(W^{1+\delta}\log(W))$, then $X^*$ and $\tilde{X}^*$ can be coupled so that they are equal with high probability. However, we will also need to make sure (in Lemma 3), that with high probability, $X^S_T$ and $X^*_T$ do not become different (and similarly for $\tilde{X}^S_T$ and $\tilde{X}^*_T$). This will allow us to conclude, in Theorem 6, that $X^S_T=X^*_T=\tilde{X}^*_T=\tilde{X}^S_T$ where $S$ is constant (because $W$ will be chosen independently of $T$).

**Lemma 3** *If $S=O(W^{1+\delta}\log(W))$, then for sufficiently large $W$, there is a conditional distribution $P(X^S,X^*|X)$ such that $P(X^S|X)$ is given by $K^S_\delta$, $P(X^*|X)$ is given by $K^S_{\delta,W}$, and, $\forall x\, P(X^*_T \neq X^S_T|X=x) < \epsilon/3$.*

**Proof:** Let $X'^{(0)}=X^{*(0)}=X$. For $s \geq 1$ pick $t_s$ according to the distribution $g_\delta$. If $t_s > T-W$, $X'^{(s-1)}=X^{*(s-1)}$, and $X'^{(s+1)}=X^{*(s+1)}$, then sample $x_{t_s}$ from $P(X_{t_s}|X'^{(s-1)}_{t_s-1},X'^{(s-1)}_{t_s+1},y_{t_s})$ and let $X'^{(s)}_{t_s}=X^{*(s)}_{t_s}=x_{t_s}$. If, instead, $t_s \leq T-W$, then only change $X'^{(s)}_t$. Finally, if $t_s > T-W$ but $X'^{(s-1)}_{\text{mb}(t_s)} \neq X^{*(s-1)}_{\text{mb}(t_s)}$, generate $X'^{(s)}_{t_s}$ and $X^{*(s)}_{t_s}$ independently. Set $X^S=X'^{(S)}$ and $X^*=X^{*(S)}$.

Let $S_0$ be the first time such that $t_{S_0}=T-W$, and for $j > 0$, let $S_j$ be the first time after $S_{j-1}$ such that $t_{S_j}=T-W+j$. Initially, $X'^{(0)}=X^{*(0)}$, and so, by definition of our coupling, the only way it could happen that $X'^{(S)}_T \neq X^{*(S)}_T$ is that $S_W \leq S$. Intuitively, for a "disagreement" to reach timestep $T$, it has to start before $T-W$ and "percolate" towards $T$, one step at a time. Each $S_j - S_{j-1}$ is a geometric variable with parameter $g(T-W+j)=\alpha_\delta(W+1-j)^{-1-\delta}$, and so

$$
\begin{aligned}
E(S_W - S_0) &= \sum_{j=1}^{W} E(S_j - S_{j-1}) \\
&= \alpha_\delta \sum_{j=1}^{W} j^{1+\delta} \propto W^{2+\delta}
\end{aligned}
$$

But by assumption, $S=O(W^{1+\delta}\log(W)) \ll O(W^{2+\delta})=E(S_W-S_0)$. So, for large enough $W$, by a Chernoff bound, it happens with probability greater than $1-\epsilon/3$ that $S_W \geq S_W - S_0 \geq S$, in which case $X^S_T=X^*_T$. ■

Combining the couplings from Corollary 2 and Lemma 3,

**Theorem 6** *For a DBN with mixing parameter $\eta$, there exists $S$ such that for all $T$ and all evidence sequences of length $T$, the decayed MCMC algorithm with polynomial decay $g_\delta$ mixes in at most $S$ steps.*

**Proof:** Let $\epsilon > 0$, and $x,\tilde{x} \in \Omega$. Pick $W$ large enough that the conclusion of Lemma 3 holds with $S=C_2(\eta)W^{1+\delta}\log(W)\log(1/\epsilon)$, so there is a distribution $P_\delta(X^S,X^*|X=x)$ such that $P_\delta(X^S|X=x)$ is given by $K^S_\delta$, $P_\delta(X^*|X=x)$ is given by $K^S_{\delta,W}$, and $P_\delta(X^S_T=X^*_T|X=x) > 1-\epsilon/3$. Next, by Corollary 2, there exists a joint distribution $P_{W,\delta}(X^*,\tilde{X}^*|X=x,\tilde{X}=\tilde{x})$ having the same marginal on $X^*$ as $P_\delta$, such that $P_{W,\delta}(X^*=\tilde{X}^*|X=x,\tilde{X}=\tilde{x}) \geq 1-\epsilon/3$. Since we have already generated $X^*$, we can generate $\tilde{X}^*$ from the *conditional* distribution $P_{W,\delta}(\tilde{X}^*|X=x,\tilde{X}=\tilde{x},X^*)$. Finally, by Lemma 3, there is a joint distribution $\tilde{P}_\delta(\tilde{X}^*,\tilde{X}^S|\tilde{X}=\tilde{x})$ with the correct marginals, such that $\tilde{P}_\delta(\tilde{X}^*_T=\tilde{X}^S_T|\tilde{X}=\tilde{x}) > 1-\epsilon/3$, and we generate $\tilde{X}^S$ from $\tilde{P}_\delta(\tilde{X}^S|\tilde{X}=\tilde{x},\tilde{X}^*)$. We have specified a distribution $P(X^S,X^*,\tilde{X}^S,\tilde{X}^*|X,\tilde{X})$ and, by a union bound, $P(X^S_T \neq \tilde{X}^S_T|X=x,\tilde{X}=\tilde{x}) < \epsilon$. Now marginalize out $X^*$ and $\tilde{X}^*$, and apply Corollary 1 to get the desired mixing time bound. ■

## 5 Empirical analysis

We now give some experimental results, on both synthetic and real-world example DBNs, performed using Kevin Murphy's toolbox [Murphy, 2001]. We first look at some simple, artificial DBNs. The advantages of doing this are that we can compute the exact posterior and therefore the error of our algorithm, and also that it is easy to precisely control the mixing parameter of such DBNs. In general, the performance of Monte Carlo approximation algorithms depends not so much on the complexity of the underlying graph as on the determinism in the transition model. So we expect that the qualitative behaviour that we observe here will carry over to larger models.

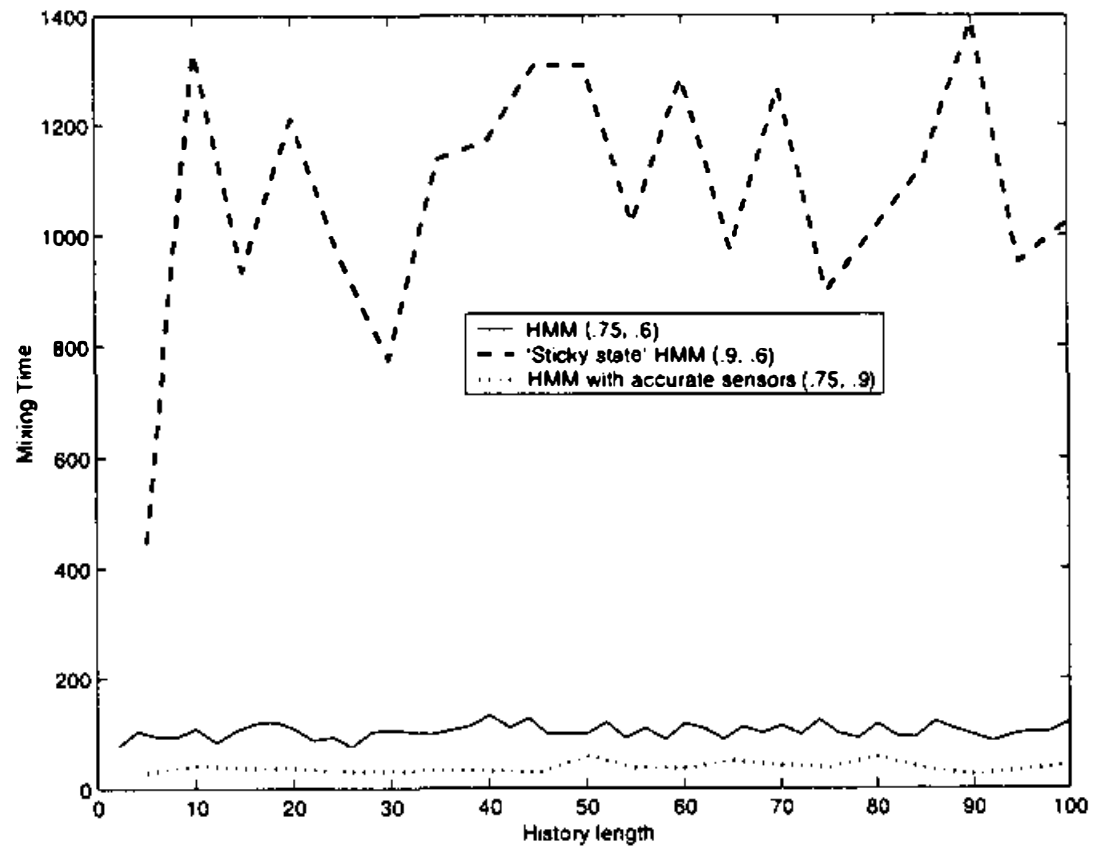

Figure 2: Mixing time ($\epsilon = .05$) as a function of history length.

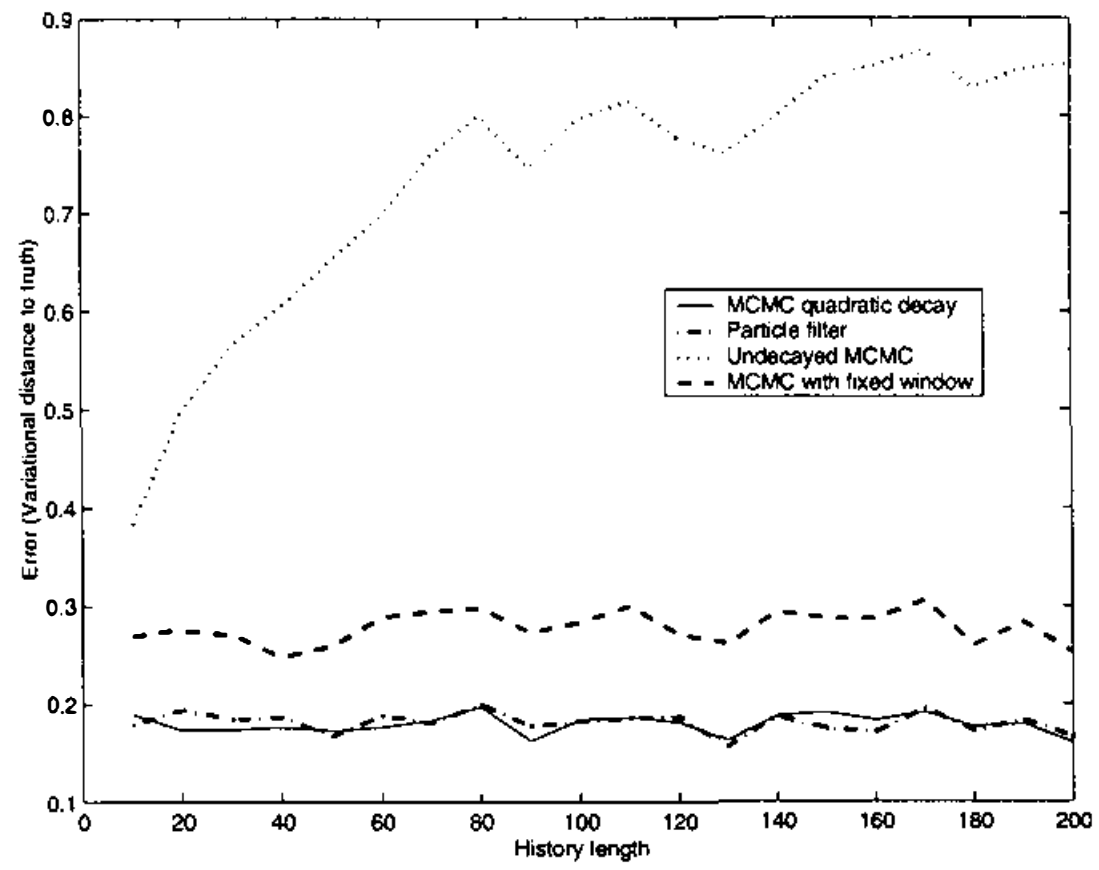

Figure 5: Error as a function of time for the WATER DBN, using 1000 samples.

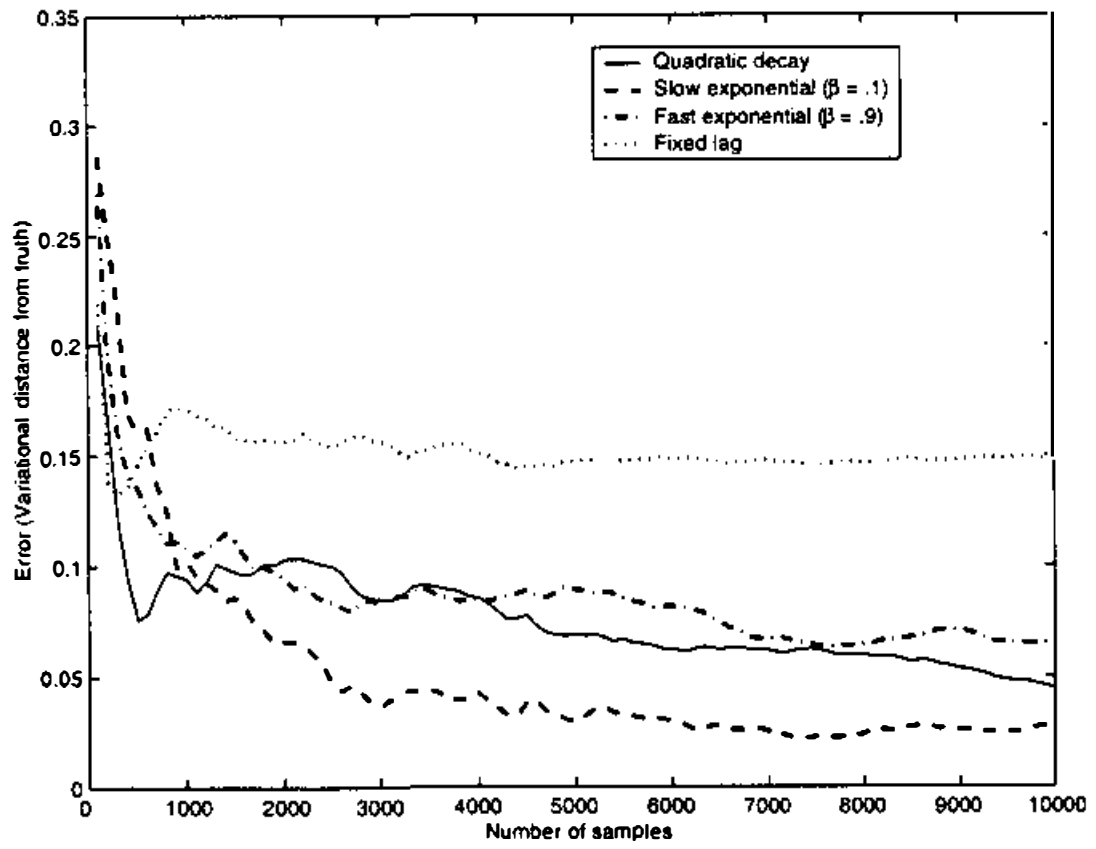

Figure 3: Error as a function of number of samples for an HMM with slow mixing parameter.

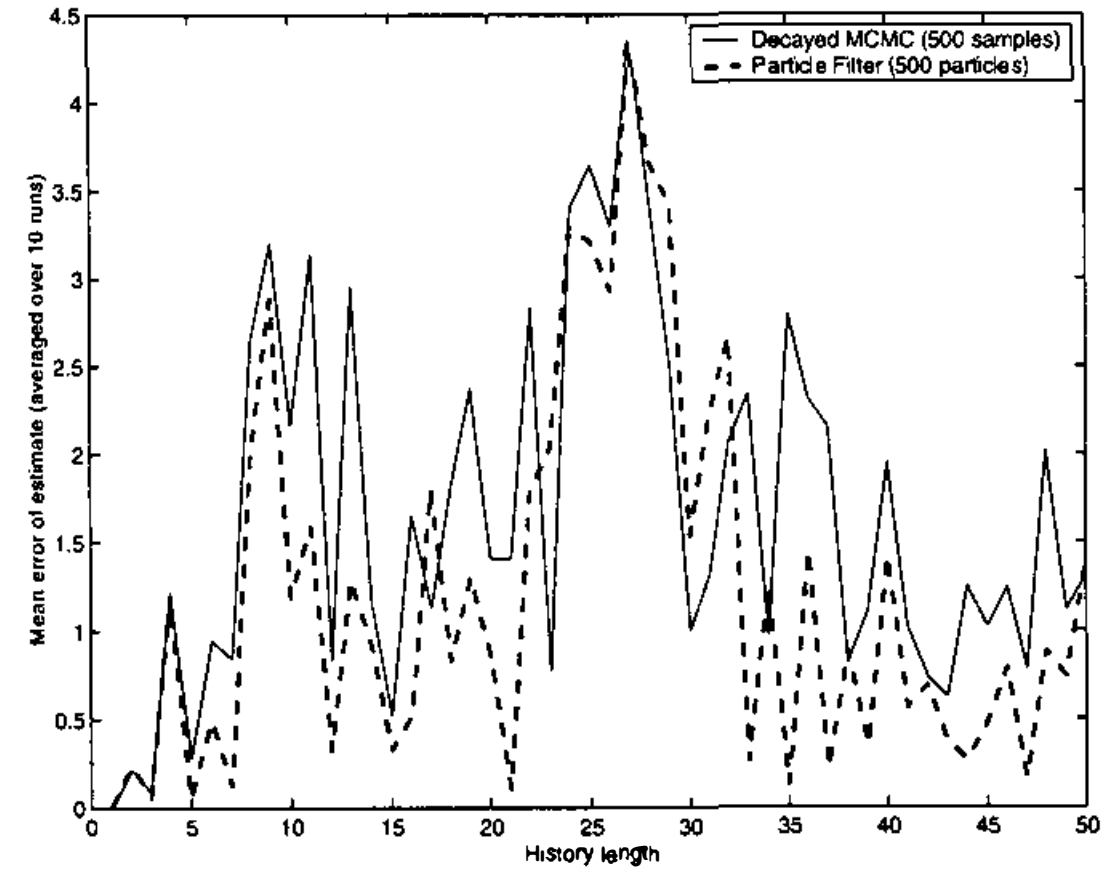

Figure 6: Error versus time for a Switching Kalman Filter, for Decayed MCMC with quadratic decay using 500 samples with gap 3, and Particle Filter with 500 particles.

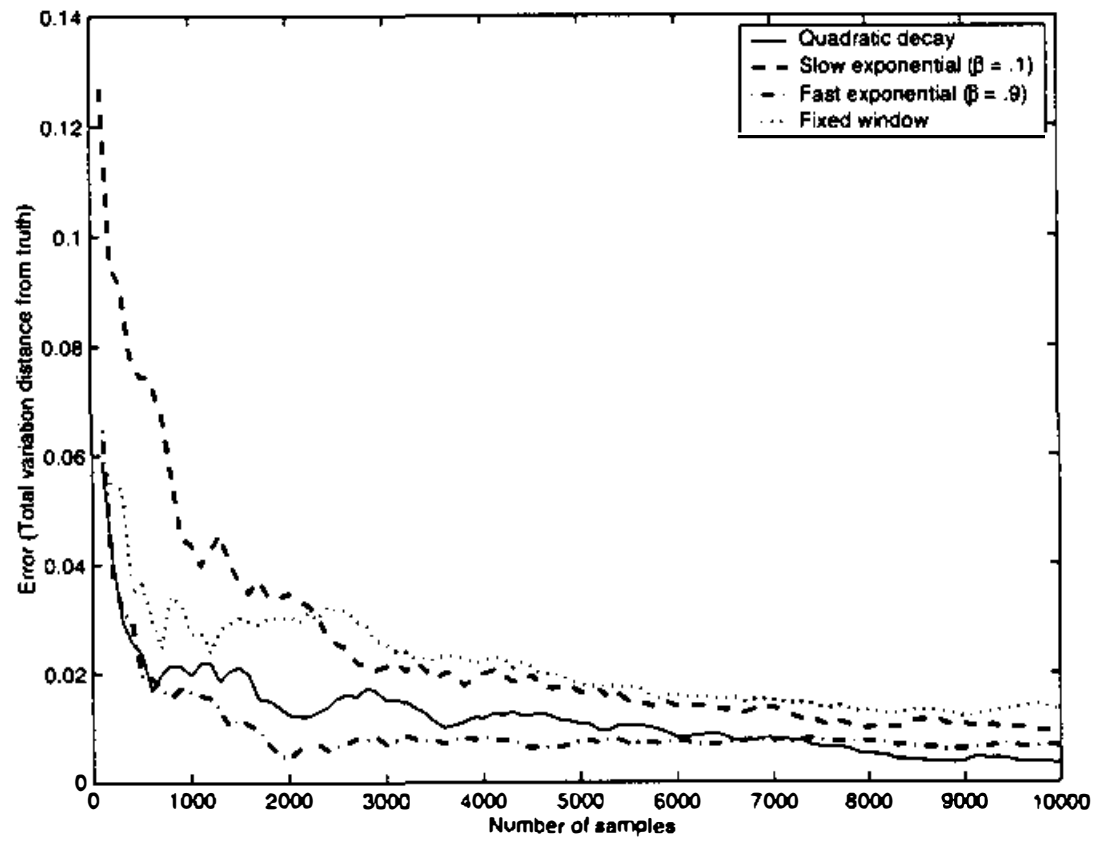

Figure 4: Error as a function of number of samples for an HMM with fast mixing parameter.

We begin by performing an experiment to verify our theorems on bounded convergence. Figure 2 shows mixing time (for a quadratic decay) as a function of history length for various DBNs with $\epsilon = .05$. As can be seen, the mixing time depends strongly on the mixing parameter of the DBN. Determinism in the transition model increases the mixing time, while determinism in the observation model decreases it (since the increasing importance of the observations means that history becomes less relevant). However, for given transition and observation models, the mixing time remains bounded as the history length increases.

The second experiment demonstrates the convergence of the various decay functions for DBNs with different mixing parameters. Figures 3 and 4 show error as a function of number of samples, for two different HMMs with fixed his-tory length 1000. The first point is that the fixed-window error converges very fast, but not to 0, since it ignores history beyond a certain point. In Figure 3, which is an HMM with a slow mixing parameter, the fast exponential decay does well initially, but then the rate of convergence slows because the decay function rarely samples beyond a certain point. The slow exponential decay performs better on this example. On the other hand, in Figure 4, the situation is reversed, and the fast exponential outperforms the slow one, because in this case it is a better match for the forgetting rate of the DBN. The quadratic decay is more robust, performing well for both HMMs.

We next consider a larger DBN – the WATER network [Jensen *et al.*, 1989], used for monitoring a water purification plant. Figure 5 shows error as a function of history length, using 1000 samples. Undecayed MCMC shows the expected increase in error, as the samples are forced to cover more ground. Among the other algorithms, fixed-window MCMC does slightly worse than the other two, because it ignores history beyond a certain point. Particle filtering and MCMC with a quadratic decay have almost identical performance. The error of decayed MCMC remains bounded, as suggested by our theoretical results.

Finally, we consider an example with continuous state for which exact inference is intractable, namely a switching Kalman filter. It consists of a switch variable $S_t$ taking finitely many values, a continuous state variable $X_t = X_{t-1} + S_t + v_t$, and observation $Y_t = X_t + w_t$, where $v_t$ and $w_t$ are Gaussian. This can model, for example, noisy observations of the position of a maneuvering object. Figure 6 shows error (measured as the distance between the mean of the samples to the true value) versus history length.

## 6 Conclusions

We have described a simple approximate filtering algorithm called *decayed MCMC*. Experimentally, it has performance comparable to other filtering algorithms. Also, being an MCMC algorithm, it is amenable to theoretical analysis, and we have shown that it comes with strong convergence guarantees.

Directions for future work include generalizing the convergence proofs to continuous state spaces and improving the algorithm using parallel chains. Another interesting possibility is to choose the number of samples adaptively based on recent evidence – an option not available with sequential methods.

### Acknowledgements

This work was supported by NSF ECS-9873474 and ONR N00014-00-1-0637. Bhaskara Marthi was also supported by an NSERC PGS-A fellowship.